\documentclass[11pt]{article}

\usepackage[final]{acl}

\usepackage{times}
\usepackage{latexsym}

\usepackage[T1]{fontenc}

\usepackage[utf8]{inputenc}

\usepackage{microtype}

\usepackage{inconsolata}

\usepackage{graphicx}

\usepackage{amsmath}
\usepackage{amssymb}
\usepackage{amsfonts}
\usepackage{algorithm}
\usepackage{algpseudocode}
\usepackage{mathtools} 

\usepackage{booktabs}   
\usepackage{multirow}   
\usepackage{adjustbox}
\usepackage[normalem]{ulem}
\useunder{\uline}{\ul}{}
\usepackage{graphicx}   
\usepackage{colortbl}   
\usepackage{xcolor}     
\definecolor{Gray}{gray}{0.9} 

\newcommand{\methodname}{MACS}

\title{\textit{\methodname}: Modality-Aware Capacity Scaling for Efficient Multimodal MoE Inference}

\author{
Bo Li \textsuperscript{1},
Chuan Wu \textsuperscript{2,3},
Shaolin Zhu \textsuperscript{2}\thanks{ Corresponding Author.}
\\
\textsuperscript{1} School of Software, Tsinghua University, Beijing, China\\
\textsuperscript{2} TJUNLP Lab, School of Computer Science and Technology, Tianjin University, China \\
\textsuperscript{3} School of New Media and Communication, Tianjin University, China
}

\begin{document}
\maketitle
\begin{abstract}
Mixture-of-Experts Multimodal Large Language Models (MoE MLLMs) suffer from a significant efficiency bottleneck during Expert Parallelism (EP) inference due to the straggler effect.
This issue is worsened in the multimodal context, as existing token-count-based load balancing methods fail to address two unique challenges: (1) Information Heterogeneity, where numerous redundant visual tokens are treated equally to semantically critical ones, and (2) Modality Dynamics, where varying visual to text ratios across tasks lead to resource misallocation.
To address these challenges, we propose \textbf{\methodname} (\textbf{M}odality-\textbf{A}ware \textbf{C}apacity \textbf{S}caling), a training-free inference framework. Specifically, \methodname~introduces an Entropy-Weighted Load mechanism to quantify the semantic value of visual tokens, addressing information heterogeneity. Additionally, the Dynamic Modality-Adaptive Capacity mechanism allocates expert resources based on the real-time modal composition of the input.
Extensive experiments demonstrate that \methodname~significantly outperforms existing methods on various multimodal benchmarks, providing a novel and robust solution for the efficient deployment of MoE MLLMs in EP inference.
\end{abstract}

\section{Introduction}

Multimodal Large Language Models (MLLMs) have demonstrated remarkable capabilities in perceiving and reasoning across diverse modalities~\cite{openai2025chatgpt,liu2024improvedbaselinesvisualinstruction,bai2025qwen3vltechnicalreport}.
To efficiently scale MLLMs, the Mixture-of-Experts (MoE) architecture has become a mainstream choice~\cite{fedus2022switch,qu2024llama,wang2025internvl3, team2025kimi, bai2025qwen3vltechnicalreport}.
By sparsely activating a subset of experts for each token,
MoE theoretically decouples the size of the model parameters from the inference computation, striking a balance between efficiency and performance~\cite{fedus2022switch}.

In practice, MoE MLLMs are often deployed using \textbf{Expert Parallelism} (EP)~\cite{cai2024shortcut}, where different experts are distributed across multiple computational devices to improve throughput. 
However, this paradigm introduces an unavoidable synchronization bottleneck: all devices must wait after processing their respective tokens until the most heavily loaded device has finished its computation before proceeding to the next layer. 
CAI-MoE~\cite{he2025capacity} formally defines this phenomenon as the straggler effect, where the overall inference latency is determined by the most heavily loaded straggler expert. 
Although this work proposes effective mitigation strategies, such as token drop, its methods are primarily designed for unimodal text models, under the core assumption that each token represents roughly equal computational load.

Recent studies indicate that the straggler effect is significantly worsened in MoE MLLMs under EP inference~\cite{li2025token,wu2025unveiling}. 
Specifically, multimodal inputs highlight two deeper sources of load imbalance:
(i) Information Heterogeneity. 
Unlike text tokens, which have a relatively uniform semantic density~\cite{li-etal-2023-mmnmt}, a single visual input is typically encoded into hundreds of patch tokens, many of which correspond to low-information background regions~\cite{liang2025explaining,wu2025unveiling}.
However, token-count-based capacity management, as used in CAI-MoE, treats redundant background tokens and semantically critical object or text tokens equally, inevitably causing severe misestimation of true computational load and resource misallocation.
(ii) Modality Dynamics. 
The ratio of visual to textual tokens varies dramatically between tasks, ranging from image-intensive document understanding or OCR tasks to text-dominant reasoning tasks. 
With such highly dynamic modality compositions, traditional token-count-based load modeling fails to accurately capture the actual computational pressure on experts, further increasing load imbalance and synchronization delays~\cite{xue2024openmoe,zhang2025mixture}.

To address these challenges, we propose \textbf{\methodname}
(\textbf{M}odality-\textbf{A}ware \textbf{C}apacity \textbf{S}caling), a training-free inference framework for MoE MLLMs.
We revisit expert capacity allocation under EP inference from a modality-aware perspective.
Specifically, we employ an Entropy-Weighted Load mechanism
to quantify and differentiate the semantic value of visual tokens, thereby mitigating load imbalance caused by information heterogeneity.
In addition, the Dynamic Modality-Adaptive Capacity mechanism
adjusts the expert capacity based on the real-time modality composition of each input batch, effectively alleviating the amplified straggler effect in multimodal settings and significantly improving the inference efficiency.
Finally, to handle inevitable capacity overflows,
we design a two-phase overflow handling mechanism to minimize information loss.

The main contributions of our work are summarized as follows:
\textbf{(I)} We systematically analyze the core mechanisms through which the straggler effect is acutely exacerbated in MoE MLLMs under EP inference, driven by visual token redundancy and modality dynamics.
\textbf{(II)} We propose \methodname, which enables more fine-grained and robust expert load scheduling at the inference stage through its Entropy-Weighted Load and Dynamic Modality-Adaptive Capacity mechanisms.
\textbf{(III)} We demonstrate through extensive experiments that \methodname~outperforms existing methods on various multimodal benchmarks, offering a novel and effective solution for the efficient deployment of MoE MLLMs in EP inference.

\begin{figure*}[t]
    
    \centering
    \includegraphics[width=1.0\textwidth]{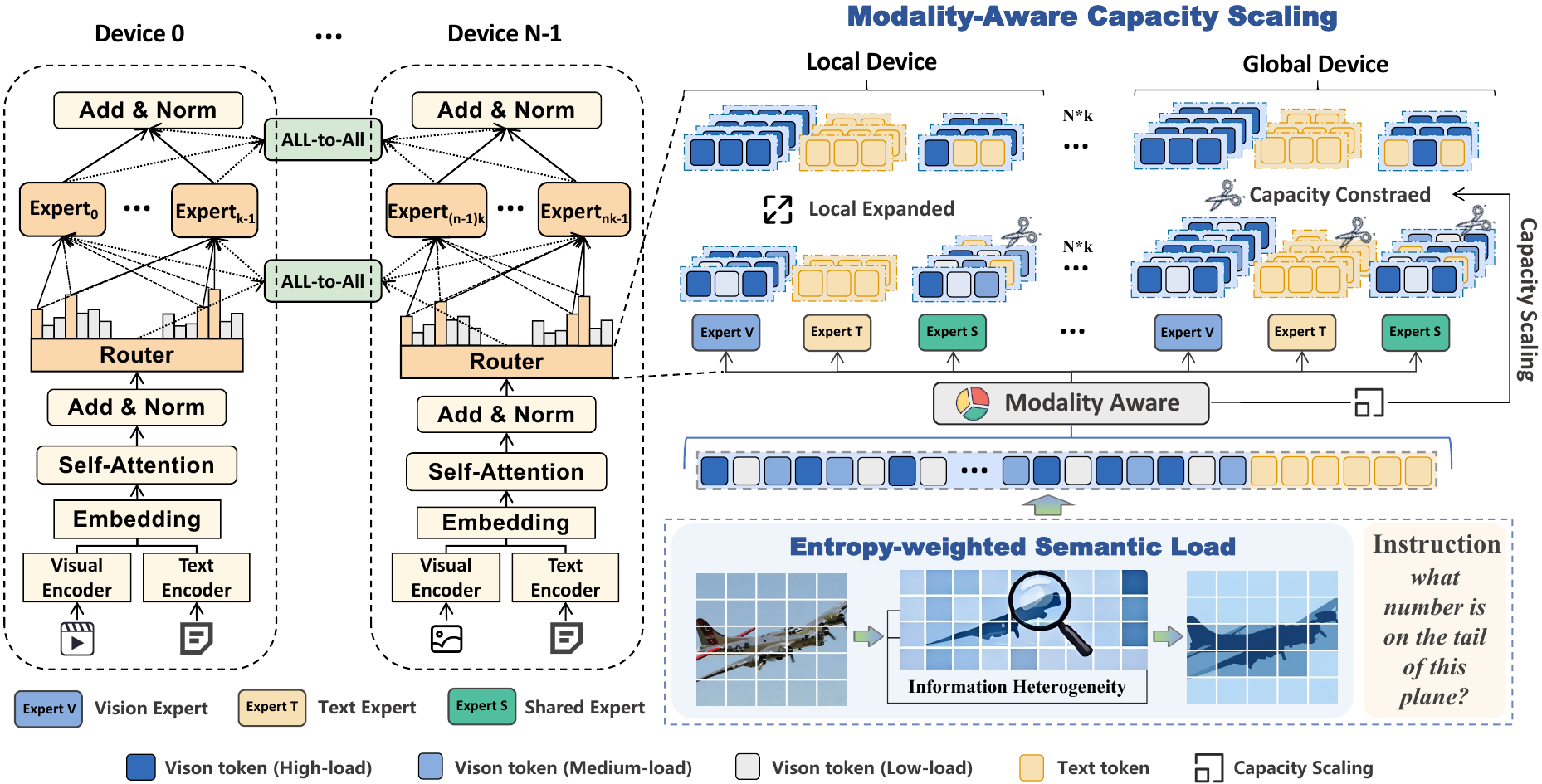}
    \caption{Overview of the \methodname~framework. It consists of three components: Entropy-Weighted Load, which models expert load based on token information; Dynamic Modality-Adaptive Capacity, which adjusts expert capacity according to batch-level modality composition, and Local Semantic Rerouting, which locally reroutes overflow tokens and applies a fail-safe drop when rerouting is infeasible.}
    \label{fig:arch}
\end{figure*}


\section{Related Work}
\label{sec:related_work}
This work addresses the efficiency bottleneck of MoE MLLMs under EP inference. 

\paragraph{MoE Models under EP.}
MoE models are often deployed using the EP distributed strategy to improve throughput~\cite{cai2024shortcut}. 
However, this approach introduces a synchronization bottleneck that leads to the straggler effect~\cite{he2025capacity}, where overall system latency is determined by the slowest expert. 
To mitigate this issue, existing research primarily falls into two categories:
(I) Capacity Management and Token Dropping.
Capacity-Aware Inference (CAI-MoE)~\cite{he2025capacity} addresses the straggler effect by imposing a capacity limit on experts and discarding excess tokens. 
While effective, its core mechanism relies on token counting, assuming all tokens have equal computational value, a premise with significant limitations in multimodal contexts.
(II) Expert Pruning and Dynamic Skipping.
Stun~\cite{lee2025stun} and MoE-Pruner~\cite{xie2024moe} reduce the computational load by decreasing the number of activated experts, including permanently removing redundant experts through structured pruning.
NAEE~\cite{lu2024not} and MC-MoE~\cite{huang2024mixture} dynamically skip non-essential experts during inference, primarily making decisions based on signals such as routing probabilities.
However, these methods are mostly designed for unimodal text models and often suffer from performance degradation when directly applied to multimodal architectures, as they cannot handle the unique behaviors of different modalities.

\paragraph{Imbalance in MoE MLLMs.}
Recent studies on MLLM interpretability have revealed that the straggler effect is acutely exacerbated in multimodal contexts, stemming from two deeper challenges:
(I) Information Heterogeneity. 
\citet{wu2025unveiling, zhang2026evaluatingsteeringmodalitypreferences} identified significant internal functional specialization. 
\citet{li2025token, liang2025explaining} have shown that multimodal inputs themselves exhibit high information heterogeneity, many visual tokens correspond only to regions with low-information background. 
For a load balancing system based solely on token counting, this intrinsic information difference is imperceptible.
(II) Modality Dynamics. 
The ratio of visual to textual tokens varies dramatically between tasks, ranging from image-intensive document understanding or OCR tasks~\cite{li-etal-2025-mit, zhu-etal-2023-peit} to text-dominant reasoning tasks~\cite{DBLP:conf/aaai/LiYJZS25, zuo-etal-2025-inimagetrans}.
With such highly dynamic modality compositions, traditional token-count-based load modeling fails to accurately capture the actual computational pressure on experts, further increasing load imbalance and synchronization delays~\cite{xue2024openmoe, zhang2025mixture, zhang-etal-2025-merge}.

Based on these observations, we propose~\textbf{\methodname}, which effectively mitigates the straggler effect under expert parallelism through its Entropy-Weighted Load and Dynamic Modality-Adaptive Capacity mechanisms.

\section{Methodology}
\label{sec:methodology}

We present \methodname, a training-free inference framework for MoE MLLMs.
As illustrated in Figure~\ref{fig:arch}, it consists of three components:
(I) Entropy-Weighted Load (Sec.~\ref{sec:entropy_load}), which models expert load based on token information;
(II) Dynamic Modality-Adaptive Capacity (Sec.~\ref{sec:dynamic_capacity}), which adjusts expert capacity
according to batch-level modality composition, and
(III) Local Semantic Rerouting (Sec.~\ref {sec:rerouting}), which locally reroutes overflow tokens and applies a fail-safe drop when rerouting is infeasible.

\subsection{Problem Formulation}
\label{sec:formulation}

A standard MoE layer consists of $N$ experts $\mathcal{E}=\{E_1,\dots,E_N\}$ and a router network $G(\cdot)$ that produces gating scores for each input token $x$. The router selects the top-$k$ experts and computes the output as
\begin{equation}
y(x)=\sum_{j\in \mathrm{Top}\text{-}k(G(x))} G(x)_j \cdot E_j(x).
\end{equation}
Let $\mathcal{T}$ denote the set of tokens in a batch and $\mathcal{I}_j\subset\mathcal{T}$ the set of tokens assigned to expert $E_j$. Under expert parallelism, the inference latency of an MoE layer, denoted as $\mathcal{L}_{\mathrm{MoE}}$, is bounded by the slowest expert due to synchronization.
\begin{equation}
\mathcal{L}_{\mathrm{MoE}} \propto \max_{j\in\{1,\dots,N\}} |\mathcal{I}_j|.
\end{equation}
The \textbf{straggler effect} arises when $\max_j |\mathcal{I}_j| \gg \mathrm{mean}_j |\mathcal{I}_j|$, creating a severe bottleneck. Existing approaches typically mitigate this issue by imposing a static capacity limit
\begin{equation}
C = \gamma \cdot \frac{|\mathcal{T}| \cdot k}{N},
\end{equation}
where $\gamma$ is a fixed capacity factor. 
However, in multimodal settings, raw token counts are a poor proxy for computational demand due to substantial information heterogeneity among tokens, particularly on the visual side.

\subsection{Entropy-Weighted Expert Load}

\label{sec:entropy_load}

To reduce redundant visual tokens that consume expert capacity, we replace the count-based load metric with an information-based one, using entropy as a proxy for semantic importance.

\paragraph{Entropy Computation and Normalization.}
For a visual token $x_v$ with a feature representation $z\in\mathbb{R}^D$, we compute its Shannon entropy $H(x_v)$ from the probability distribution obtained by $\mathrm{Softmax}(z)$. To ensure robustness across different images and models, we apply image-wise z-score normalization over visual tokens:
\begin{equation}
\tilde{H}(x_v)=\frac{H(x_v)-\mu_{\mathcal{B}}}{\sigma_{\mathcal{B}}+\epsilon},
\end{equation}
where $\mu_{\mathcal{B}}$ and $\sigma_{\mathcal{B}}$ denote the mean and standard deviation of the entropy values within the current batch $\mathcal{B}$, and $\epsilon$ is a small constant for numerical stability.

\paragraph{Semantic Weighting and Effective Load.}
We define a semantic weight function
\begin{equation}
w(x)=
\begin{cases}
\sigma\!\left(-\delta \cdot \tilde{H}(x)\right), & x\in\mathcal{T}_{vis},\\
1.0, & x\in\mathcal{T}_{txt},
\end{cases}
\end{equation}
where $\sigma(\cdot)$ is the Sigmoid function and $\delta$ controls the sensitivity of the entropy-to-weight mapping. Text tokens are assigned full weight due to their high semantic density. 
The effective load of the expert $E_j$ is then defined as
\begin{equation}
\tilde{L}_j = \sum_{x\in\mathcal{I}_j} w(x),
\end{equation}
allowing experts to process a larger number of low-information visual tokens without prematurely reaching capacity limits.

\subsection{Modality-Aware Capacity Scaling}

\label{sec:dynamic_capacity}

A static capacity factor is agnostic to the modality composition of the input batch. To prevent expert overload in vision-heavy scenarios and resource underutilization in text-heavy ones, we dynamically scale expert capacities based on the batch's effective modality ratio.

\paragraph{Effective Modality Ratio.}
Using semantic weights, we compute the effective visual ratio
\begin{equation}
R_v = \frac{\sum_{x\in\mathcal{T}_{vis}} w(x)}{\sum_{x\in\mathcal{T}} w(x)},
\end{equation}
which better reflects the true computational demand of the visual modality than raw token proportions.

\paragraph{Adaptive Capacity Scaling.}
Following prior analyses of expert specialization, we categorize experts into three groups based on their activation frequencies on a held-out calibration set: visual experts $\mathcal{E}_{vis}$, text experts $\mathcal{E}_{txt}$ and shared experts $\mathcal{E}_{shared}$. We define a modality bias indicator
\begin{equation}
m_j=
\begin{cases}
+1, & E_j\in\mathcal{E}_{vis},\\
-1, & E_j\in\mathcal{E}_{txt},\\
0, & E_j\in\mathcal{E}_{shared}.
\end{cases}
\end{equation}
Let $C_{base}$ denote the base capacity derived from the static formulation. We scale the capacity of each expert as
\begin{equation}
C_j = C_{base} \cdot \left(1 + \rho \cdot m_j \cdot (R_v - 0.5)\right),
\end{equation}
where $\rho$ controls the adaptation strength. In practice, we clamp $C_j$ to a minimum value to avoid degenerate capacities. When $R_v>0.5$, visual experts receive increased capacity while text experts are constrained, and vise versa.

\begin{table*}[t]
\centering
\resizebox{\textwidth}{!}{%
\setlength{\tabcolsep}{2.0pt} 
\begin{tabular}{l| ccccccc ccccc |c} 
\toprule
\multirow{2}{*}{\textbf{Method}} & \multicolumn{7}{c}{\textbf{Image Understanding}} & \multicolumn{5}{c}{\textbf{Video Understanding}} & \textbf{Avg.} \\
\cmidrule(lr){2-8} \cmidrule(lr){9-13}
 & TextVQA & ChartQA & MMStar & MMBench & MMVet & MME & RWQA & MVBench & EgoSch & VMME & LVB & VMMMU & (\%) \\
\midrule
\multicolumn{14}{c}{\textsc{Qwen3-VL-30B-A3B-Instruct} ($\gamma_0=1.0$)} \\
\midrule
Vanilla MoE 
& 83.54 & 85.36 & 72.10 & 86.82 & 85.67 & 2500 & 73.72 
& 72.29 & 63.28 & 74.53 & 62.47 & 68.64 & 100.00 \\

CAI-MoE (Token Drop) 
& 77.42 & 79.13 & 66.58 & 81.34 & 79.21 & 2214 & 67.89 
& 66.41 & 58.12 & 68.34 & 56.89 & 62.17 & 91.80 \\

CAI-MoE (Expanded) 
& 79.86 & 81.57 & 68.42 & 83.05 & 81.63 & 2305 & 69.46 
& 68.74 & 60.15 & 70.92 & 58.73 & 64.28 & 94.69 \\

MACS (w/o Expanded) 
& 83.04 & 84.89 & 71.48 & 86.12 & 85.03 & 2478 & 73.15 
& 71.63 & 62.84 & 73.91 & 61.88 & 67.92 & 99.20 \\

\rowcolor{Gray} MACS (Ours) 
& \textbf{83.41} & \textbf{85.22} & \textbf{71.93} & \textbf{86.67} & \textbf{85.48} & \textbf{2492} & \textbf{73.58} 
& \textbf{72.11} & \textbf{63.14} & \textbf{74.36} & \textbf{62.33} & \textbf{68.49} & \textbf{99.78} \\

\midrule
\multicolumn{14}{c}{\textsc{InternVL3.5-30B-A3B} ($\gamma_0=1.0$)} \\
\midrule
Vanilla MoE 
& 85.68 & 84.14 & 72.03 & 84.68 & 85.43 & 2324 & 64.87 
& 72.06 & 60.37 & 68.65 & 63.76 & 65.24 & 100.00 \\

CAI-MoE (Token Drop) 
& 78.13 & 76.52 & 65.47 & 78.23 & 77.19 & 2056 & 58.34 
& 64.21 & 54.12 & 61.88 & 57.24 & 58.63 & 90.22 \\

CAI-MoE (Expanded) 
& 80.94 & 79.28 & 67.91 & 80.76 & 80.14 & 2147 & 61.05 
& 67.33 & 56.49 & 64.52 & 59.81 & 61.17 & 93.93 \\

MACS (w/o Expanded) 
& 84.92 & 83.47 & 71.36 & 84.02 & 84.66 & 2298 & 64.12 
& 71.14 & 59.83 & 67.94 & 63.11 & 64.58 & 98.96 \\

\rowcolor{Gray} MACS (Ours) 
& \textbf{85.51} & \textbf{83.98} & \textbf{71.84} & \textbf{84.49} & \textbf{85.17} & \textbf{2315} & \textbf{64.69} 
& \textbf{71.89} & \textbf{60.21} & \textbf{68.42} & \textbf{63.55} & \textbf{65.03} & \textbf{99.72} \\

\midrule
\multicolumn{14}{c}{\textsc{Kimi-VL-A3B-Instruct} ($\gamma_0=1.0$)} \\
\midrule
Vanilla MoE 
& 88.39 & 87.26 & 61.25 & 83.11 & 77.84 & 2218 & 68.07 
& 62.73 & 78.32 & 66.84 & 64.37 & 57.58 & 100.00 \\

CAI-MoE (Token Drop) 
& 82.56 & 81.04 & 56.83 & 77.45 & 71.27 & 2013 & 63.42 
& 57.18 & 72.56 & 61.29 & 59.14 & 52.87 & 92.24 \\

CAI-MoE (Expanded) 
& 84.72 & 83.91 & 58.17 & 79.62 & 73.84 & 2096 & 65.18 
& 59.42 & 74.89 & 63.56 & 61.22 & 54.63 & 95.28 \\

MACS (w/o Expanded) 
& 87.94 & 86.68 & 60.89 & 82.76 & 77.12 & 2198 & 67.63 
& 62.15 & 77.84 & 66.21 & 63.95 & 57.12 & 99.28 \\

\rowcolor{Gray} MACS (Ours) 
& \textbf{88.27} & \textbf{87.14} & \textbf{61.16} & \textbf{83.02} & \textbf{77.63} & \textbf{2212} & \textbf{67.96} 
& \textbf{62.58} & \textbf{78.19} & \textbf{66.67} & \textbf{64.24} & \textbf{57.49} & \textbf{99.81} \\
\bottomrule
\end{tabular}%
}
\caption{Performance comparison of \methodname~against the SOTA distributed MoE inference acceleration method CAI-MoE on multimodal benchmarks. We evaluate on Qwen3-VL, InternVL3.5, and Kimi-VL, comparing against CAI-MoE's \textit{Token Drop} and \textit{Expanded Drop} variants. ``Vanilla MoE'' denotes the unconstrained baseline. All acceleration methods use a base capacity factor $\gamma_0=1.0$. ``w/o Expanded'' denotes the variant without local expansion, while ``Ours'' represents the full method.}
\label{tbl:main_results}
\end{table*}

\subsection{Local Semantic Rerouting}
\label{sec:rerouting}

Even with information aware load modeling and adaptive capacity scaling, transient expert overflows may still occur.
When an expert $E_j$ exceeds its capacity $C_j$,
we first attempt to reroute overflow tokens locally to avoid
unnecessary token dropping and cross-device communication.

Let $\mathcal{E}_{cand}$ denote the set of experts on the same computational device whose effective loads satisfy $\tilde{L}_k < C_k$.
For an overflow token $x$ with feature representation $z_x$,
we score each candidate expert $E_k \in \mathcal{E}_{cand}$ by combining router preference and semantic affinity:
\begin{equation}
S(x, E_k) = (1-\eta)\, G(x)_k + \eta \cdot \mathrm{sim}(z_x, \mu_k),
\end{equation}
where $\mu_k$ is the semantic centroid of the expert $E_k$ and
$\mathrm{sim}(\cdot,\cdot)$ denotes the cosine similarity.
The overflow token is rerouted to the candidate expert with the highest score that satisfies the capacity constraint.

If no candidate experts are available on the local device ($\mathcal{E}_{cand} = \emptyset$), rerouting is infeasible.
In this case, we activate a fail-safe drop mechanism.
We define a retention score for an overflow token $x$ as
\begin{equation}
r(x) = w(x) \cdot \max_j G(x)_j,
\end{equation}
which jointly considers the token's semantic importance and routing confidence.
When dropping is unavoidable, tokens with the lowest retention scores
are discarded first, ensuring that only tokens of low-importance and low-confidence are removed.

\section{Experiments}

\subsection{Setup}
\paragraph{Models and Implementation.} 
We conducted experiments on three SOTA MoE MLLMs: 
Qwen3-VL (30B-A3B)~\cite{bai2025qwen3vltechnicalreport} 
, InternVL3.5 (30B-A3B)~\cite{wang2025internvl3}.
, and Kimi-VL (16B-A3B)~\cite{team2025kimi} 
These models employ distinct MoE configurations (e.g., Qwen and InternVL use 48 layers with 128 experts, whereas Kimi-VL adopts a hybrid architecture with shared experts and 64 routing experts). 
All experiments were implemented using DeepSpeed.
We performed distributed inference on 8 NVIDIA A100 GPUs, employing 8-way EP to simulate a high-performance production environment.

\paragraph{Datasets.} 
We evaluated \methodname~on a comprehensive multimodal benchmarks. 
For Image Understanding, we utilize 8 Zero-Shot benchmarks, including $\text{TextVQA}_{\text{val}}$ \cite{singh2019towards} and ChartQA \cite{masry2022chartqa},
$\text{MMBench}_{\text{en}}$ \cite{liu2024mmbenchmultimodalmodelallaround}, MMStar \cite{chen2024rightwayevaluatinglarge}, MMVet \cite{yu2023mmvet}, MME \cite{fu2023mme} and RealWorldQA \cite{realworldQA}.
For Video Understanding, we extend our evaluation to dynamic visual tasks using MVBench \cite{li2024mvbenchcomprehensivemultimodalvideo}, EgoSchema \cite{mangalam2023egoschemadiagnosticbenchmarklongform}, VideoMME \cite{fu2025video}, $\text{LongVideoBench}_{\text{val}}$ \cite{wu2024longvideobenchbenchmarklongcontextinterleaved}, and VideoMMMU \cite{hu2025videommmuevaluatingknowledgeacquisition}.
Performance is reported using standard accuracy metrics, while efficiency is measured via End-to-End Latency and Speedup.

\paragraph{Baselines.} 
We compared our approach with the original model and the SOTA distributed MoE inference acceleration method, CAI-MoE~\cite{he2025capacity}. 
Vanilla MoE serves as both the performance upper bound and the latency lower bound. 
For CAI-MoE, we evaluate both its Token Drop and Expanded Drop variants.

\begin{table}[t]
    \centering    
    \resizebox{\linewidth}{!}{
        \setlength{\tabcolsep}{3.5pt} 
        \begin{tabular}{lcccccccc}
            \toprule
            \textbf{Method} & \textbf{SC} & \textbf{SL} & \textbf{DC} & \textbf{SR} & \textbf{TextVQA} & \textbf{RWQA} & \textbf{MMB} & \textbf{VMMMU} \\
            \midrule
            \multicolumn{9}{c}{$\gamma_0 = 0.5, \quad \rho = 0.6$} \\
            \midrule
            Baseline & $\checkmark$ & - & - & - & 68.91 & 63.21 & 74.29 & 54.64 \\
            + SL     & $\checkmark$ & $\checkmark$ & - & - & 74.96 & 66.19 & 77.94 & 60.85 \\
            + DC     & - & $\checkmark$ & $\checkmark$ & - & 79.11 & 69.74 & 83.41 & 64.29 \\
            \textbf{\methodname~ (Ours)} & - & $\checkmark$ & $\checkmark$ & $\checkmark$ & \textbf{81.04} & \textbf{71.28} & \textbf{84.21} & \textbf{67.17} \\
            \midrule
            \multicolumn{9}{c}{$\gamma_0 = 1.0, \quad \rho = 0.6$} \\ 
            \midrule
            Baseline & $\checkmark$ & - & - & - & 77.42 & 67.89 & 81.34 & 62.17 \\
            + SL     & $\checkmark$ & $\checkmark$ & - & - & 79.28 & 69.96 & 82.41 & 64.37 \\
            + DC     & - & $\checkmark$ & $\checkmark$ & - & 83.04 & 73.15 & 86.12 & 67.92 \\
            \textbf{\methodname~ (Ours)} & - & $\checkmark$ & $\checkmark$ & $\checkmark$ & \textbf{83.41} & \textbf{73.58} & \textbf{86.67} & \textbf{68.49} \\
            \bottomrule
        \end{tabular}
    }
    \caption{Ablation study on Qwen3-VL. ``SC'' denotes the baseline with only Static Capacity and conventional token counting. ``+SL'' indicates the addition of Entropy-weighted Semantic Load. ``+DC'' further incorporates Modality-aware Dynamic Capacity. Experiments are conducted with two base capacity factors ($\gamma_0=0.5$ and $\gamma_0=1.0$) to evaluate performance under varying pressure levels.}
\label{tab:ablation}
\end{table}

\subsection{Main Results}

To comprehensively evaluate the effectiveness of \methodname,
we compare it with a representative capacity-aware MoE inference
acceleration method, CAI-MoE~\cite{he2025capacity},
across three mainstream MoE MLLMs.
All methods are evaluated under an EP inference setting,
with a unified base capacity factor of $\gamma_0 = 1.0$
to ensure a fair comparison.

As shown in Table~\ref{tbl:main_results}, the results demonstrate that \methodname~consistently outperforms both variants of CAI-MoE on all three models (Qwen3-VL, InternVL3.5, and Kimi-VL).
Specifically, the full \methodname~preserves more than 99.7\% of the
original Vanilla MoE performance, corresponding to an average degradation of less than 0.3\%.
In contrast, CAI-MoE incurs substantial performance losses:
its Token Drop variant degrades performance by over 7\%, while the Expanded Drop variant still results in an approximately 5\% drop.
These results suggest a fundamental limitation of token-count-based
capacity management, which inevitably discards semantically valuable tokens when experts are overloaded.
By contrast, \methodname~makes information-aware allocation decisions
that better preserve multimodal reasoning capability.

Importantly, this comparison isolates the contribution of our core design.
Even \methodname~(w/o Expanded), which excludes local semantic rerouting, significantly outperforms CAI-MoE.
For instance, on the Kimi-VL model, \methodname~(w/o Expanded) improves performance from 92.24\% (CAI-MoE Token Drop, which also lacks rerouting) to 99.28\% through Entropy-Weighted Load and
Dynamic Modality-Adaptive Capacity alone.
This observation indicates that information-aware load balancing, rather than token manipulation, is the primary factor in mitigating
performance degradation under EP inference.
Building upon this foundation, the full \methodname further improves performance from 99.28\% to 99.81\% by incorporating Local Semantic Rerouting, which effectively recovers overflowed tokens and nearly closes the performance gap with the original Vanilla MoE.


\begin{figure}[t]
\centering
\includegraphics[scale=0.56]{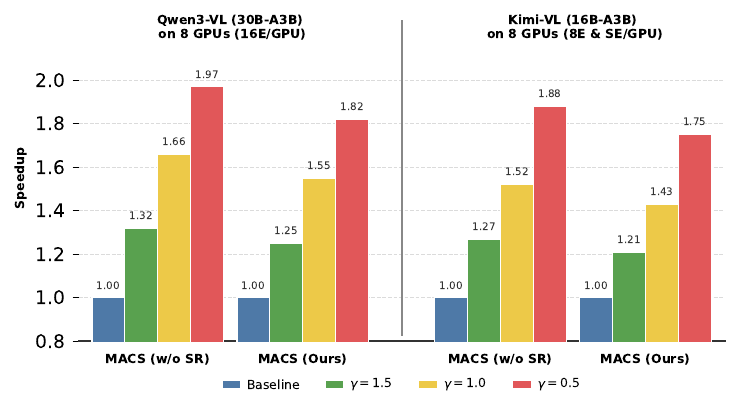}
\caption{Compared to the capacity-unconstrained baseline, the speedup of a single MoE layer is achieved through two capacity-aware inference methods: Capacity Constrained and Semantic Rerouting.}
\label{fig:Inference Speedup}
\end{figure}
\begin{figure}[t]
\centering
\includegraphics[scale=0.6]{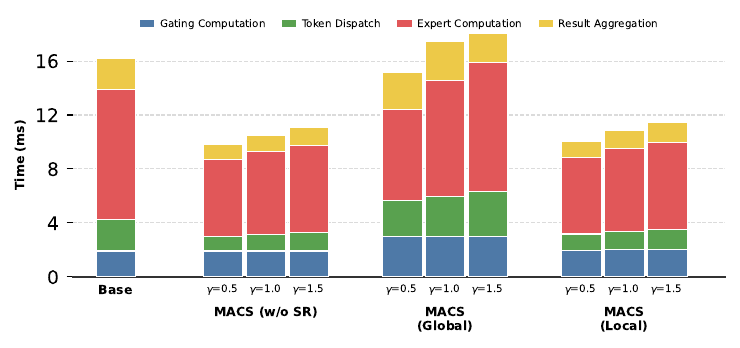}
\caption{Inference latency speedup across different stages of Qwen3-VL. Global refers to rerouting overflowed tokens within the global expert scope, while Local denotes rerouting overflowed tokens exclusively to local experts.}
\label{fig:Latency Breakdown}
\end{figure}

\subsection{Ablation Study}
Table \ref{tab:ablation} presents an ablation study on Qwen3-VL under two base capacity settings, $\gamma_0 = 0.5$ and $\gamma_0 = 1.0$, where we progressively introduce its core components. 
We start from a static-capacity baseline (SC) and incrementally add Semantic Load (SL), Dynamic Modality-Adaptive Capacity (DC), and Semantic Rerouting (SR) to analyze their individual contributions.

We can find that SL consistently improves performance in both capacity settings. 
Compared to the static-capacity baseline, SL enables the model to distinguish between high-information and low-information visual tokens, reducing unnecessary capacity consumption by redundant background tokens. 
This results demonstrate that more accurate load modeling alone can effectively alleviate performance degradation under expert-parallel inference.
Building on SL, the addition of DC yields further consistent improvements. 
DC dynamically adjusts expert capacity based on the effective modality composition of each input batch, leading to more stable performance across tasks with varying visual-to-text ratios. 
This effect is particularly pronounced in the constrained capacity setting ($\gamma_0 = 0.5$), indicating that dynamic capacity allocation is especially beneficial when expert resources are limited, and load imbalance is more severe.
Incorporating SR provides additional and stable gains across all benchmarks and capacity settings. 
Although SR is not the primary source of improvement, it consistently narrows the remaining performance gap to the full-capacity model, complementing SL and DC to improve robustness under extreme load conditions.

Overall, the ablation results indicate that SL and DC are the core performance drivers of \methodname, while Semantic Rerouting serves as a lightweight yet effective auxiliary mechanism to handle unavoidable overflows in EP inference.


\subsection{Efficiency Analysis}

\begin{figure}[t]
    \centering
    \includegraphics[width=1.0\linewidth]{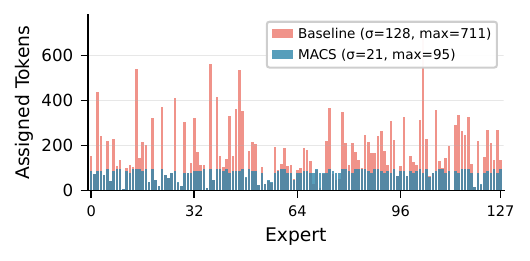}
    \caption{Mitigating the Straggler Effect in Qwen3-VL ($\gamma_0=0.5$). The x-axis shows the expert index, and the y-axis shows the expert load.}
    \label{fig:straggler_effect}
\end{figure}
As shown in Figure~\ref{fig:straggler_effect}, our method significantly mitigates the straggler effect, reducing the maximum load from 711 to 95, thereby achieving more efficient expert parallel inference. To quantify the inference acceleration provided by \methodname, we evaluated the end-to-end inference speed of the MoE layer and analyzed the sources of these gains through a latency breakdown. 
All experiments are conducted under an EP setting.

\paragraph{End-to-End Speedup.}
As shown in Figure~\ref{fig:Inference Speedup}, \methodname significantly improves the inference speed of each MoE layer compared to the unconstrained baseline. 
This speedup is consistently observed on Qwen3-VL and Kimi-VL. 
As the capacity factor $\gamma$ decreases, the inference speedup continues to increase, reaching up to 1.97$\times$ on Qwen3-VL. 
Furthermore, our Local Semantic Rerouting reassigns overflow tokens to local idle experts, and introduces negligible computational overhead while yielding substantial performance recovery (as shown in Table~\ref{tbl:main_results}), achieving both efficiency and performance.

\paragraph{Latency Breakdown Analysis.}
To analyze the components of these speedups, we decompose the MoE layer's latency, as illustrated in Figure~\ref{fig:Latency Breakdown}. 
The most significant latency reduction consistently occurs in the Expert Computation stage. 
For example, at $\gamma=0.5$, expert computation latency is reduced by over 40\% compared to the unconstrained baseline. 
This demonstrates that our capacity constraint mechanism effectively limits the number of tokens assigned to the busiest experts, reducing the system's waiting time.
This analysis also highlights the superiority of our Local Semantic Rerouting. 
In \methodname~(Global), the overhead of the Token Dispatch and Result Aggregation stage increases noticeably. 
This is because global rerouting requires broadcasting and synchronizing more token information across all GPUs. 
In contrast, the communication overhead of our \methodname~(Local) is nearly identical to that of the \methodname~(w/o SR), demonstrating the significant advantage of the local strategy in controlling communication costs.

\begin{table*}[t]
\centering
\small
\begin{adjustbox}{max width=\textwidth}
\begin{tabular}{@{}l|c|cccccccc|c@{}}
\toprule
\bf Method & \bf $\gamma$ & \bf  TextVQA &\bf   ChartQA &\bf   MMStar & \bf  MMBench &\bf   MMVet & \bf   MME & \bf  RealWorldQA & \bf  COCO & \begin{tabular}[c]{@{}c@{}} \bf Avg. (\%)\end{tabular} \\
\midrule
Baseline& $+\infty$ & 83.54 & 85.36 & 72.10 & 86.82 & 85.67 & 2500 & 73.72 & 80.28 & 100.00 \\
\midrule
Random & \multirow{4}{*}{1.5}
& 80.89 & 82.67 & 69.83 & 84.11 & 82.97 & 2422 & 71.41 & 78.92 & 97.04 \\
Router-based & 
& 81.67 & 83.48 & 70.49 & 84.92 & 83.78 & 2446 & 72.09 & 78.97 & 97.87 \\
Modality-Prior &
& 82.59 & 84.37 & 71.27 & 85.83 & 84.66 & 2471 & 72.87 & 79.47 & 98.86 \\
\bf Entropy-weighted Load(Ours) &
& \textbf{82.93} & \textbf{84.72} & \textbf{71.53} & \textbf{86.17} & \textbf{84.97} & \textbf{2481} & \textbf{73.18} & \textbf{79.52} & \underline{\textbf{99.22}} \\
\midrule
Random & \multirow{4}{*}{1.0}
& 69.94 & 71.53 & 60.36 & 72.76 & 71.74 & 2092 & 61.69 & 68.71 & 83.97 \\
Router-based &
& 73.61 & 75.19 & 63.53 & 76.49 & 75.46 & 2204 & 64.93 & 71.69 & 88.25 \\
Modality-Prior &
& 77.42 & 79.13 & 66.58 & 81.34 & 79.21 & 2214 & 67.89 & 72.36 & 91.83 \\
\bf Entropy-weighted Load(Ours) &
& \textbf{79.28} & \textbf{81.01} & \textbf{68.39} & \textbf{82.41} & \textbf{81.32} & \textbf{2372} & \textbf{69.96} & \textbf{76.21} & \underline{\textbf{94.90}} \\
\midrule
Random & \multirow{4}{*}{0.5}
& 53.79 & 54.98 & 46.43 & 55.91 & 55.17 & 1611 & 47.47 & 50.47 & 64.21 \\
Router-based &
& 61.88 & 63.26 & 53.41 & 64.32 & 63.47 & 1852 & 54.61 & 59.53 & 74.09 \\
Modality-Prior &
& 68.91 & 68.79 & 59.81 & 74.29 & 68.79 & 2007 & 63.21 & 61.98 & 81.89 \\
\bf Entropy-weighted Load(Ours) &
& \textbf{74.96} & \textbf{76.62} & \textbf{64.71} & \textbf{77.94} & \textbf{76.88} & \textbf{2246} & \textbf{66.19} & \textbf{72.37} & \underline{\textbf{89.82}} \\
\bottomrule
\end{tabular}
\end{adjustbox}
\caption{Comparison of different token selection strategies under varying capacity factors $\gamma$.
We evaluate Random, Router-based, Modality-Prior,
and \methodname~(Entropy-Weighted Load).
The baseline operates without capacity constraints ($\gamma=+\infty$).
$\gamma$ controls the severity of capacity pressure during EP inference.}
\label{tab:semantic_load_analysis}
\end{table*}

\begin{figure}[t]
    \centering
    \includegraphics[scale=0.54]{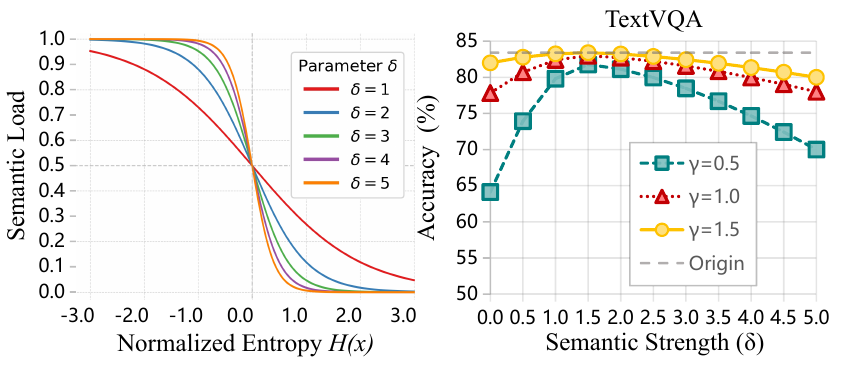}
    \caption{Sensitivity analysis of Semantic Strength ($\delta$). (Left) Entropy-to-load mapping curves. (Right) TextVQA performance trends under varying base capacities ($\gamma$).}
    \label{fig:sensitivity_delta}
\end{figure}

\begin{figure}[t]
    \centering
    \includegraphics[scale=0.54]{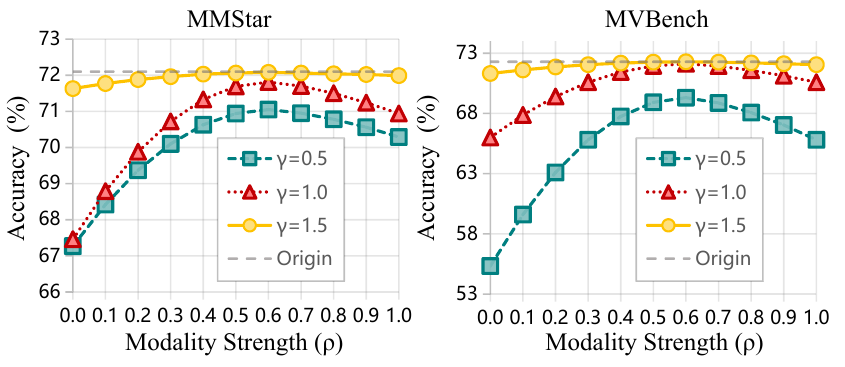}
    \caption{Sensitivity analysis of Modality Adaptation Strength ($\rho$). Performance on MMStar (image) and MVBench (video) under varying base capacities ($\gamma_0$). The gray dashed line denotes the unconstrained upper bound.}
    \label{fig:sensitivity_rho}
\end{figure}

\subsection{Parameters Sensitivity Analysis}

As illustrated in Figure~\ref{fig:sensitivity_delta}, $\delta$ governs the non-linearity of the entropy mapping. 
Specifically, a low $\delta$ ($<1.0$) results in a flattened weight distribution that fails to effectively suppress background noise. 
Experimental results on TextVQA demonstrate that $\delta \approx 1.5$ strikes the optimal balance, particularly in low-capacity scenarios ($\gamma=0.5$), as it effectively distinguishes between foreground and background information without compromising subtle yet critical visual details (e.g., small OCR characters).

We investigated the impact of the modality adaptation strength $\rho$ on multimodal tasks under varying base capacities ($\gamma$), as illustrated in Figure~\ref{fig:sensitivity_rho}. 
Specifically, for image tasks (MMStar), increasing $\rho$ significantly overcomes the static bottleneck in resource-constrained scenarios ($\gamma=0.5$), while under the standard setting ($\gamma=1.0$), setting $\rho=0.6$ restores the model to its original performance level. 
It is worth noting that when $\rho > 0.8$, a ``Cannibalization Effect'' is observed, where the excessive capacity expansion of visual experts encroaches upon the resources required for text reasoning, leading to a performance regression. 
In the context of video tasks (MVBench) with high-frame-rate inputs, a static low-capacity configuration ($\gamma=0.5$) results in a performance collapse.
However, \methodname~achieves a significant recovery of +12.5\%, demonstrating its critical capability in handling highly redundant visual streams. 
In conclusion, $\rho=0.6$ exhibits the best robustness across various modalities and resource constraints, achieving an effective balance between visual throughput and textual reasoning.

\subsection{Analysis of Entropy-Weighted Load}
\label{sec:semantic_load_analysis}

To evaluate the effectiveness of Entropy-Weighted Load for quantifying token-level information value and expert load regulation, we conduct experiments on Qwen3-VL by comparing token dropping strategies based on three alternative weighting criteria: (1) Random, which discards tokens uniformly at random; (2) Router-based, which prioritizes tokens according to routing confidence; and (3) Modality-Prior, which preferentially drops visual tokens assumed to be more redundant.
We vary the capacity factor $\gamma \in \{1.5, 1.0, 0.5\}$ to simulate different levels of capacity constraints under EP inference.

As shown in Table~\ref{tab:semantic_load_analysis}, our Entropy-weighted Load consistently achieves the best performance under all capacity settings. 
When $\gamma=0.5$, the Entropy-weighted Load still preserves $89.82\%$ of the baseline performance, while all alternative strategies show substantial degradation.
The Random strategy performs the worst across all settings, indicating that capacity reduction alone, without information-aware token prioritization, is insufficient to maintain model performance. 
The Router-based strategy improves over random selection, but its routing scores primarily reflect an expert's preference for individual tokens rather than the tokens' intrinsic semantic importance. 
As a result, semantically critical tokens with low routing confidence may be erroneously discarded. 
The Modality-Prior strategy improves performance on language-sensitive tasks by enforcing the retention of text tokens, but it applies a coarse-grained approach to visual tokens. 
By failing to distinguish informative foreground regions from redundant background content, it leads to notable performance degradation on vision-centric tasks such as COCO.

These results demonstrate that the Entropy-Weighted Load  provides an effective measure of token priority by enabling fine-grained modeling of the density of cross-modal information. 
This property allows for more reliable token retention and dropping decisions under capacity-constrained EP inference, and forms a solid foundation for the subsequent dynamic capacity scaling and semantic rerouting mechanisms in \methodname.

\section{Conclusion}
In this work, we identify a significant efficiency bottleneck in MoE MLLMs under EP inference, where the straggler effect is worsened by two challenges unique to the multimodal domain: the information heterogeneity of visual tokens and modality dynamics across tasks.
To address these challenges, we propose \methodname, which mitigates information heterogeneity through an Entropy-Weighted Load mechanism and adapts to modality dynamics with its Dynamic Modality-Adaptive Capacity. Extensive experiments demonstrate that MACS significantly outperforms existing methods across various MoE MLLMs, providing a practical solution for the efficient deployment of multimodal MoE models.


\bibliography{custom}
\clearpage
\appendix

\section{Appendix}
\label{sec:appendix}

\subsection{Implementation Details}
\paragraph{Hyperparameter Settings.}
For the main experiments, we employ a consistent hyperparameter configuration for the MACS framework, based on the sensitivity analysis presented in Section 4.5. 

Specifically, the semantic strength parameter $\delta$ in the Entropy-Weighted Load mechanism is set to $1.5$. This value was selected to reduce the weights of low-information background tokens while retaining necessary visual details (e.g., small OCR characters). 

The modality adaptation strength $\rho$ in the Dynamic Modality-Adaptive Capacity mechanism is set to $0.6$. This setting balances visual throughput with the capacity required for textual reasoning, which helps mitigate resource competition between modalities (referred to as the ``Cannibalization Effect''). Unless otherwise specified (e.g., in ablation studies), these parameters ($\delta=1.5, \rho=0.6$) are applied consistently across all evaluated models (Qwen3-VL, InternVL3.5, and Kimi-VL) and benchmarks.

\subsection{Expert Calibration and Classification Details}
\label{sec:appendix_calibration}

\paragraph{Calibration Set Configuration}
To mitigate distribution bias and ensure statistical significance during expert classification, we construct a balanced calibration dataset $\mathcal{D}_{calib} = \mathcal{D}_{txt} \cup \mathcal{D}_{vis}$ comprising a total of 16,384 samples. The dataset is strictly stratified into two modalities:

\begin{itemize}
    \item \textbf{Text Modality} ($\mathcal{D}_{txt}$, $N=8192$): Randomly sampled from the MMLU benchmark\cite{hendryckstest2021}. This covers a broad spectrum of domains, including STEM and humanities, ensuring the generality of the text expert activation distribution.
    \item \textbf{Visual Modality} ($\mathcal{D}_{vis}$, $N=8192$): Randomly sampled from the LLaVA-OneVision dataset\cite{li2024llava}, encompassing general imagery alongside complex visual scenarios such as OCR and charts.
\end{itemize}

Regarding convergence, our empirical observations demonstrate that the Kullback-Leibler (KL) divergence of the expert activation distribution reaches convergence ($\Delta KL < 10^{-3}$) at $N \approx 6000$. Consequently, allocating 8,192 samples per modality provides a sufficient margin to accurately capture the intrinsic routing preferences of the experts.

\paragraph{Classification Basis}
To quantify the modality bias of each expert, we first compute the activation frequency $f_{j}^{(m)}$ of expert $E_j$ for a given modality $m \in \{txt, vis\}$:
\begin{equation}
    f_{j}^{(m)} = \frac{1}{|\mathcal{D}_m|} \sum_{x \in \mathcal{D}_m} \mathbb{I}\big(E_j \in \text{TopK}(G(x))\big)
\end{equation}
where $\mathbb{I}(\cdot)$ represents the indicator function and $G(x)$ denotes the routing network. We subsequently define the modality specialization score $\Delta_{j}$ for each expert as the difference in activation frequencies:
\begin{equation}
    \Delta_{j} = f_{j}^{(vis)} - f_{j}^{(txt)}
\end{equation}

Based on this score, we partition the expert set $\mathcal{E}$ into three mutually exclusive subsets using a threshold $\delta = 0.1$. Experts exhibiting a significant modality preference are categorized as \textit{modality-specific experts}: visual experts are defined as $\mathcal{E}_{vis} = \{ E_j \in \mathcal{E} \mid \Delta_{j} \ge \delta \}$, and text experts as $\mathcal{E}_{txt} = \{ E_j \in \mathcal{E} \mid \Delta_{j} \le -\delta \}$. Conversely, experts demonstrating minimal variance in activation frequency across modalities are designated as \textit{multimodal shared experts}, defined as $\mathcal{E}_{shared} = \{ E_j \in \mathcal{E} \mid |\Delta_{j}| < \delta \}$. These shared experts are primarily responsible for cross-modal alignment and general reasoning tasks.

\subsection{Centroid Computation and Memory Overhead}
\label{sec:appendix_centroid}

\paragraph{Offline Centroid Computation}
The computation of expert centroids is performed offline prior to model deployment, introducing zero training overhead. To obtain stable representations, we reuse the aforementioned calibration dataset. For each expert $E_k$ within a given layer, we aggregate the hidden states $z$ of all tokens routed to it. The centroid $\mu_k$ is derived via mean pooling of these token embeddings:
\begin{equation}
    \mu_k = \frac{1}{|Z_k|} \sum_{z \in Z_k} z
\end{equation}
where $Z_k$ represents the set of tokens assigned to expert $E_k$. This procedure is completed in advance, and the computed centroids $\mu_k$ remain static and frozen throughout the entire inference phase.

\paragraph{Memory Overhead Analysis}
The memory footprint required to store these static centroids is negligible relative to the overall VRAM usage of the large language model. The storage overhead $M$ is calculated as $M = L \times N \times D \times P$, where $L$ denotes the total number of layers, $N$ is the number of experts per layer, $D$ represents the hidden dimension size, and $P$ indicates the byte size of the numerical precision (e.g., $P=2$ for FP16 precision).

Taking the Qwen3-VL-30B architecture as an illustrative example ($L=48$, $N=128$, $D=2048$), the total memory overhead is approximately $48 \times 128 \times 2048 \times 2$ bytes, which equals roughly $25$ MB. Given that the parameter size of the 30B model requires approximately 60 GB of memory, this additional centroid storage constitutes less than $0.05\%$ of the total memory footprint. Consequently, the proposed mechanism introduces minimal impact on practical deployment resource constraints.

\begin{figure}[t]
    \centering
    \includegraphics[width=0.94\linewidth]{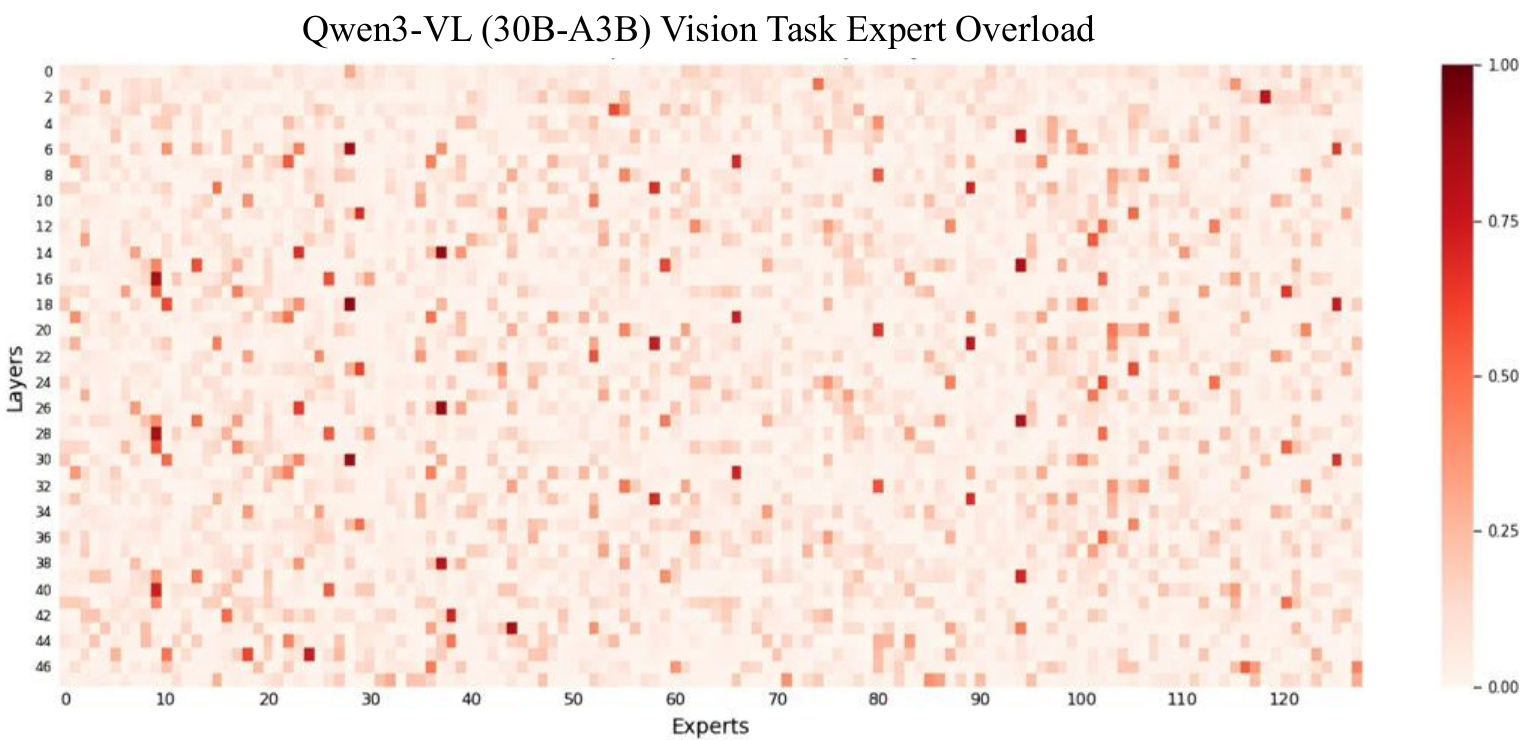}
    \includegraphics[width=0.94\linewidth]{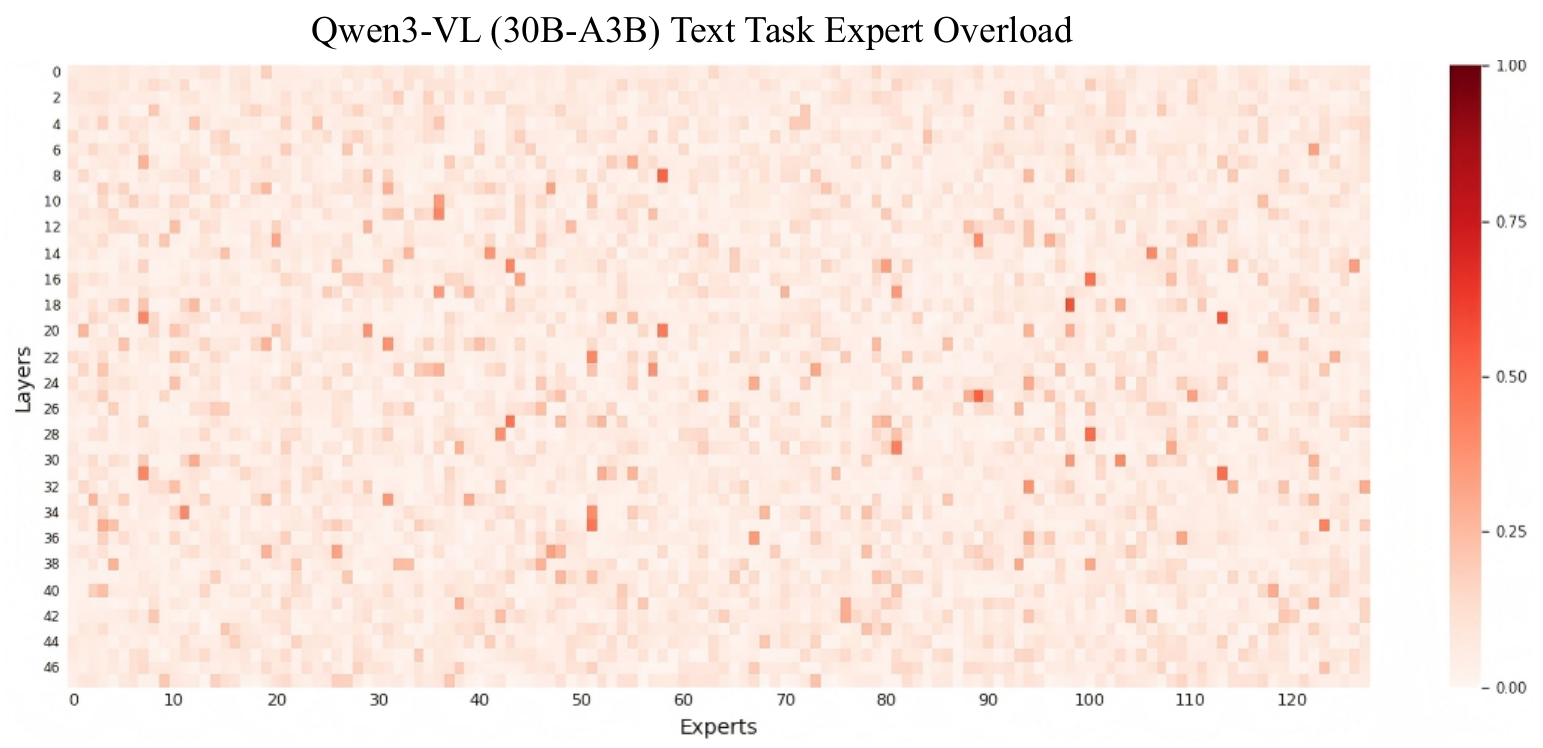}
    \caption{Normalized expert load on vision and text tasks in Qwen3-VL-30B-A3B.}
    \label{fig:heatmap}
\end{figure}

\subsection{Straggler Effect in Multimodal Scenarios}

We further analyze the normalized expert load heatmaps presented in Figure~\ref{fig:heatmap} to investigate the root cause of inference latency in multimodal MoE models.

\paragraph{Disparity in Load Distribution.}
As illustrated in Figure~\ref{fig:heatmap}, there is a sharp contrast between the activation patterns of vision (top) and text (bottom) tasks. The text modality exhibits a relatively uniform and sparse distribution of expert utilization. In contrast, the vision task demonstrates significant load skewness, characterized by distinct ``hotspots'' (dark red clusters) where the normalized load approaches 1.0. This indicates that visual tokens tend to disproportionately congregate on specific experts, rapidly saturating their capacity.

\paragraph{Exacerbated Straggler Effect.}
The observed load imbalance in the vision modality directly contributes to a severe Straggler Effect. In MoE inference, the latency of a layer is bounded by the expert with the highest computational load (the straggler). The dense high-load clusters in the vision heatmap suggest that visual tasks create higher synchronization barriers compared to text tasks. Consequently, in a unified architecture, the vision modality acts as the primary bottleneck, intensifying the latency tail and justifying the necessity for modality-aware capacity management strategies.

\begin{table*}[t]
\centering
\resizebox{\textwidth}{!}{%
\setlength{\tabcolsep}{2.0pt} 
\begin{tabular}{l| ccccccc ccccc |c} 
\toprule
\multirow{2}{*}{\textbf{Method}} & \multicolumn{7}{c}{\textbf{Image Understanding}} & \multicolumn{5}{c}{\textbf{Video Understanding}} & \textbf{Avg.} \\
\cmidrule(lr){2-8} \cmidrule(lr){9-13}
 & TextVQA & ChartQA & MMStar & MMBench & MMVet & MME & RWQA & MVBench & EgoSch & VMME & LVB & VMMMU & (\%) \\
\midrule
\multicolumn{14}{c}{\textsc{Qwen3-VL-30B-A3B-Instruct} ($\gamma_0=0.5$)} \\
\midrule
Vanilla MoE 
& 83.54 & 85.36 & 72.10 & 86.82 & 85.67 & 2500 & 73.72 
& 72.29 & 63.28 & 74.53 & 62.47 & 68.64 & 100.00 \\

CAI-MoE (Token Drop) 
& 68.91 & 68.79 & 59.81 & 74.29 & 68.79 & 2007 & 63.21 
& 60.22 & 50.37 & 61.58 & 49.73 & 54.64 & 81.89 \\

CAI-MoE (Expanded) 
& 71.51 & 68.47 & 58.18 & 72.58 & 70.66 & 2144 & 60.99 
& 58.89 & 56.11 & 62.93 & 53.21 & 55.91 & 83.52 \\

MACS (w/o Expanded) 
& 79.11 & \textbf{81.66} & 67.61 & 83.41 & 81.04 & 2376 & 69.74 
& \textbf{71.06} & \textbf{60.42} & 69.99 & \textbf{60.43} & 64.29 & 95.21 \\

\rowcolor{Gray} MACS (Ours) 
& \textbf{81.04} & 80.96 & \textbf{68.84} & \textbf{84.21} & \textbf{83.56} & \textbf{2424} & \textbf{71.28} 
& 69.76 & 60.32 & \textbf{71.61} & 60.18 & \textbf{67.17} & \textbf{96.47} \\

\midrule
\multicolumn{14}{c}{\textsc{InternVL3.5-30B-A3B} ($\gamma_0=0.5$)} \\
\midrule
Vanilla MoE 
& 85.68 & 84.14 & 72.03 & 84.68 & 85.43 & 2324 & 64.87 
& 72.06 & 60.37 & 68.65 & 63.76 & 65.24 & 100.00 \\

CAI-MoE (Token Drop) 
& 71.14 & 64.54 & 59.76 & 68.46 & 68.32 & 1948 & 55.21 
& 61.11 & 47.98 & 55.66 & 52.92 & 55.41 & 82.14 \\

CAI-MoE (Expanded) 
& 72.66 & 71.97 & 59.96 & 70.29 & 75.21 & 2077 & 55.67 
& 63.78 & 52.46 & 57.93 & 55.41 & 58.61 & 86.36 \\

MACS (w/o Expanded) 
& 82.21 & \textbf{82.76} & 66.32 & 82.34 & 82.12 & 2220 & 62.36 
& 67.03 & 57.76 & 66.27 & \textbf{62.62} & 62.12 & 95.84 \\

\rowcolor{Gray} MACS (Ours) 
& \textbf{82.67} & 81.46 & \textbf{70.74} & \textbf{82.67} & \textbf{82.67} & \textbf{2273} & \textbf{63.18} 
& \textbf{70.81} & \textbf{58.32} & \textbf{66.57} & 61.79 & \textbf{62.52} & \textbf{97.14} \\

\midrule
\multicolumn{14}{c}{\textsc{Kimi-VL-A3B-Instruct} ($\gamma_0=0.5$)} \\
\midrule
Vanilla MoE 
& 88.39 & 87.26 & 61.25 & 83.11 & 77.84 & 2218 & 68.07 
& 62.73 & 78.32 & 66.84 & 64.37 & 57.58 & 100.00 \\

CAI-MoE (Token Drop) 
& 71.76 & 70.72 & 49.01 & 65.71 & 57.87 & 1737 & 53.52 
& 48.17 & 62.16 & 54.56 & 56.36 & 46.47 & 79.89 \\

CAI-MoE (Expanded) 
& 75.37 & 73.54 & 48.23 & 70.16 & 65.92 & 2038 & 57.13 
& 53.57 & 66.11 & 54.14 & 56.61 & 49.96 & 84.89 \\

MACS (w/o Expanded) 
& \textbf{85.73} & 82.41 & \textbf{59.97} & 77.87 & \textbf{75.26} & \textbf{2197} & 64.21 
& 59.56 & 75.16 & 63.53 & 60.17 & 55.22 & 95.70 \\

\rowcolor{Gray} MACS (Ours) 
& 84.89 & \textbf{85.14} & 58.91 & \textbf{81.99} & 74.97 & 2146 & \textbf{66.66} 
& \textbf{60.14} & \textbf{76.23} & \textbf{65.78} & \textbf{61.47} & \textbf{56.02} & \textbf{96.99} \\
\bottomrule
\end{tabular}
}
\caption{Performance comparison of \methodname~against the SOTA distributed MoE inference acceleration method CAI-MoE on multimodal benchmarks. We evaluate on Qwen3-VL, InternVL3.5, and Kimi-VL, comparing against CAI-MoE's \textit{Token Drop} and \textit{Expanded Drop} variants. ``Vanilla MoE'' denotes the unconstrained baseline. All acceleration methods use a base capacity factor $\gamma_0=0.5$. ``w/o Expanded'' denotes the variant without local expansion, while ``Ours'' represents the full method.}
\label{tbl:main_results_0.5}
\end{table*}

\subsection{Capacity-Tradeoff Analysis}

\begin{figure}[t]
    \centering
    \includegraphics[width=0.94\linewidth]{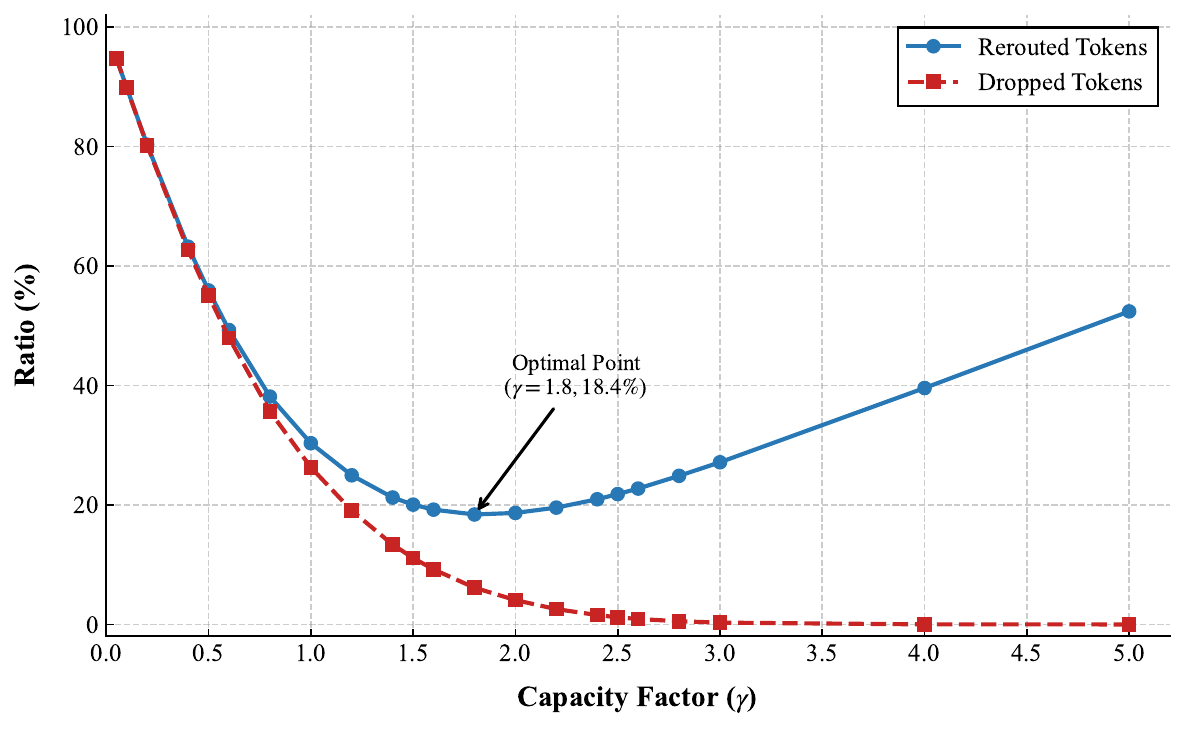}
    \caption{Impact of Capacity Factor ($\gamma_0$) on Token Rerouting and Dropping Rates.}
    \label{fig:cap_anal}
\end{figure}

As illustrated in Figure~\ref{fig:cap_anal}, evaluating the base capacity factor $\gamma_0$ from $0.05$ to $5.0$ reveals a non-linear relationship between capacity allocation and routing behavior. We categorize this into three operational regimes:

\paragraph{Congestion Regime ($\gamma_0 < 1.5$).}
In this lower capacity range, both token drop and rerouting rates are high. The limited expert capacity requires the router to either discard tokens or redirect them to alternative experts. As $\gamma_0$ increases toward $1.5$, the drop rate exhibits a continuous decrease, indicating that adding capacity directly mitigates token loss.

\paragraph{Optimal Efficiency Regime ($\gamma_0 \approx 1.8$).}
At $\gamma_0 = 1.8$, the system reaches an optimal operational point. The rerouting rate achieves its global minimum at $18.4\%$, and the drop rate is simultaneously reduced to a low level. This configuration provides an effective balance between preserving token information and maintaining stable routing assignments.

\paragraph{Saturation and Expansion Regime ($\gamma_0 > 2.0$).}
When $\gamma_0 > 2.0$, the drop rate stabilizes near zero, while the rerouting rate exhibits a increase. This rise in rerouting is driven by our local expert expansion strategy. As capacity becomes abundant, the routing network actively utilizes the available slots in idle experts to process tokens not originally assigned to them. This saturated expansion mechanism ensures load balancing and increases the utilization of experts with low loads. Although this process may cause certain tokens to be processed by more than $k$ experts, maintaining a strict $k$-expert constraint per token is unnecessary. Permitting the selection of additional experts improves the model's representational capacity, whereas enforcing a strict limit would introduce redundant computation.

\subsection{Performance  under Low Capacity Constraints}

We present a performance comparison of MACS against the inference acceleration method, CAI-MoE, across three multimodal MoE architectures: Qwen3-VL, InternVL3.5, and Kimi-VL. As detailed in Table~\ref{tbl:main_results_0.5}, all acceleration methods are evaluated under a constrained capacity setting of $\gamma_0=0.5$.

\paragraph{Performance against Baseline Methods.}
Under this constrained setting, existing methods such as CAI-MoE (Token Drop) experience performance decreases, retaining $81.89\%$ of the Vanilla MoE performance on average for Qwen3-VL. This suggests that non-selective token dropping strategies may discard necessary visual information. MACS, by contrast, maintains higher performance levels, achieving $96.47\%$ on Qwen3-VL and $97.14\%$ on InternVL3.5. This indicates that modality-aware capacity allocation helps preserve model performance under tight capacity constraints.

\paragraph{Evaluation Across Modalities.}
The performance retention of MACS is observed across both image understanding (e.g., TextVQA, MMBench) and video understanding benchmarks (e.g., MVBench). For instance, on the ChartQA benchmark, the performance of Qwen3-VL drops from $85.36$ to $68.79$ when using CAI-MoE (Token Drop). MACS mitigates this decrease, achieving a score of $80.96$. This suggests the proposed framework is more effective at retaining tokens necessary for these reasoning tasks.

\paragraph{Contribution of Local Expansion.}
The results in Table~\ref{tbl:main_results_0.5} also illustrate the effect of the local expansion mechanism. Comparing \textit{MACS (w/o Expanded)} with the full \textit{MACS (Ours)}, there is a consistent performance improvement across the evaluated models. For example, on Qwen3-VL, the average performance increases from $95.21\%$ to $96.47\%$, representing a $+1.26\%$ relative improvement. While the modality-aware capacity scaling provides the primary performance retention, the local expansion mechanism further utilizes available expert capacity to process additional tokens, contributing to overall performance without introducing significant communication overhead.

\end{document}